\documentclass[11pt]{article}

\usepackage[preprint]{acl}

\usepackage{microtype}

\usepackage[english,bidi=default]{babel}
\babelprovide[import]{hindi}
\babelprovide[import]{arabic}

\usepackage{graphicx}
\usepackage{subcaption}
\usepackage{tcolorbox}
\tcbuselibrary{breakable}
\usepackage{amsmath}
\usepackage{amsthm}
\usepackage{amssymb}
\usepackage{mathtools}
\usepackage{empheq}
\usepackage{bm}
\usepackage{tabularray}
\usepackage{caption}
\usepackage{enumitem}
\usepackage{color} %
\usepackage{url}
\usepackage{booktabs}
\usepackage{multirow}
\usepackage{rotating}
\usepackage{gb4e}
\noautomath

\renewcommand{\footnotesize}{\small}

\title{Analysis of the Neglect-Zero Effect in Large Language Models}

\author{Jin Tanaka$^{1,2}$, Daiki Matsuoka$^{1,2}$, Ryoma Kumon$^{1,2}$, Hitomi Yanaka$^{1,2,3}$ \\
  $^1$The University of Tokyo, $^2$RIKEN, $^3$Tohoku University \\
  \texttt{\{jt141214, daiki.matsuoka, kumoryo9, hyanaka\}@is.s.u-tokyo.ac.jp} \\}

\begin{document}

\maketitle

\begin{abstract}
    We investigate the extent to which the language processing of LLMs resembles human cognitive processes, focusing on a human cognitive bias called the \textit{neglect-zero effect}.
    This effect refers to the human tendency to ignore \textit{zero-models}, which are configurations that render a proposition vacuously true by virtue of an empty set.
    We focus on two types of inferences driven by the neglect-zero effect, and examine how LLMs process these inferences by comparing their behavior with that in an inference that does not involve the neglect-zero effect.
    For this purpose, we employ a paradigm based on \textit{structural priming}, where recent exposure to a preceding sentence (the \textit{prime}) facilitates the processing of a subsequent sentence (the \textit{target}) due to their structural similarity.
    We prepare primes to force LLMs to consider the zero-model, and analyze whether they also consider it in the target.
    The results suggest that the neglect-zero effect may not occur in the LLMs analyzed in this study.
    Our code is available at \url{https://github.com/ynklab/neglect_zero}.
\end{abstract}

\section{Introduction}\label{chap: introduction}

In recent years, the performance of large language models (LLMs) has improved dramatically, and their way of processing language superficially resembles that of humans to a significant degree.
However, the mechanisms underlying their language processing remain unclear, and there is growing interest in how similar LLMs' language processing is to that of humans~\citep{niu2025largelanguagemodelscognitive}.
Motivated by this interest, this study focuses on the \textit{neglect-zero effect}~\citep{Aloni-2022}, a human cognitive bias in language processing.
This effect is the human tendency to ignore \textit{zero-models}, which are configurations where a proposition is rendered vacuously true due to an empty set.

It has been hypothesized that the neglect-zero effect underlies several linguistic phenomena that are difficult to explain within conventional theories~\citep{Aloni-2022}, and this hypothesis has already been experimentally verified with human subjects~\citep{klochowicz2025neglect}.
However, it remains underexplored whether LLMs also exhibit this effect.
This study aims to address this question by applying an experimental procedure used for humans to LLMs.

Specifically, we compare the behavior of LLMs in two types of inferences, which are considered to stem from the neglect-zero effect, against their behavior in an inference called \textit{scalar implicature}, where the neglect-zero effect does not occur.
For this verification, we utilize \textit{structural priming}, a psychological phenomenon in humans.
Structural priming refers to the tendency for the processing strategy of a natural language sentence (\textit{the target}) to become similar to that of a preceding sentence (\textit{the prime}) when the two have the same syntactic or semantic structures.
Since structural priming arises from shared structures across sentences, leveraging this cognitive tendency enables us to investigate whether a common underlying mechanism drives the processing of the inference in the prime and the target.
Given that the existence of structural priming has been experimentally demonstrated in LLMs regarding other linguistic phenomena~\citep{jumelet-etal-2024-language}, appropriately adapting the methodology of~\citet{klochowicz2025neglect} into prompts is expected to elucidate whether the neglect-zero effect also exists in LLMs.

The overall results of our experiment provide negative evidence for the existence of the neglect-zero effect in LLMs.
In particular, we find that the Gemma-3 series and GPT-5 nano tend not to exhibit the neglect-zero effect in the inference employed in our research.
Conversely, Gemma-3-27B and Llama-4-Scout exhibit a certain degree of sensitivity to zero-models, but in a manner different from humans.

\section{Background}\label{chap: background}

\subsection{Neglect-Zero Effect}\label{sec: neglect-zero}

In this study, we focus on two types of inferences that are considered to be driven by the neglect-zero effect: \textit{non-empty-scope strengthening in superlative quantifiers} (ESQ) and \textit{distributive inference} (DIS).
We provide an example of each inference below.

\begin{exe}
    \ex\label{ex: ESQ} ESQ
    \begin{xlist}
        \ex\label{ex: ESQ_pre} \{Fewer than three / At most two\} circles are red.
        \ex\label{ex: ESQ_ante} $\leadsto$ There are some red circles.
    \end{xlist}
    \ex\label{ex: DIS} DIS
    \begin{xlist}
        \ex\label{ex: DIS_pre} Each circle is red or blue.
        \ex\label{ex: DIS_ante} $\leadsto$ There is a red circle and a blue circle.
    \end{xlist}
\end{exe}

\noindent
It is worth emphasizing that these inferences are not instances of logical entailment, in that they are not derived from the literal meaning of the sentences.
Regarding (\ref{ex: ESQ}), the premise (\ref{ex: ESQ_pre}) literally means that the number of red circles in the set is less than three.
Since this does not rule out the possibility that no red circles exist, the truth of (\ref{ex: ESQ_pre}) does not logically guarantee the truth of (\ref{ex: ESQ_ante}).
The same holds for (\ref{ex: DIS}).
Given that the premise (\ref{ex: DIS_pre}) literally means that every circle is colored either red or blue, it is true even in situations where all circles are red or all circles are blue.
Thus, the conclusion (\ref{ex: DIS_ante}) is not necessarily true when (\ref{ex: DIS_pre}) is true.
Therefore, (\ref{ex: ESQ}) and (\ref{ex: DIS}) are not derived from the literal meaning alone, but rather involve a certain additional factor.

\begin{figure}
    \centering
    \begin{minipage}{.43\linewidth}
        \centering
        \includegraphics[width=0.5\linewidth]{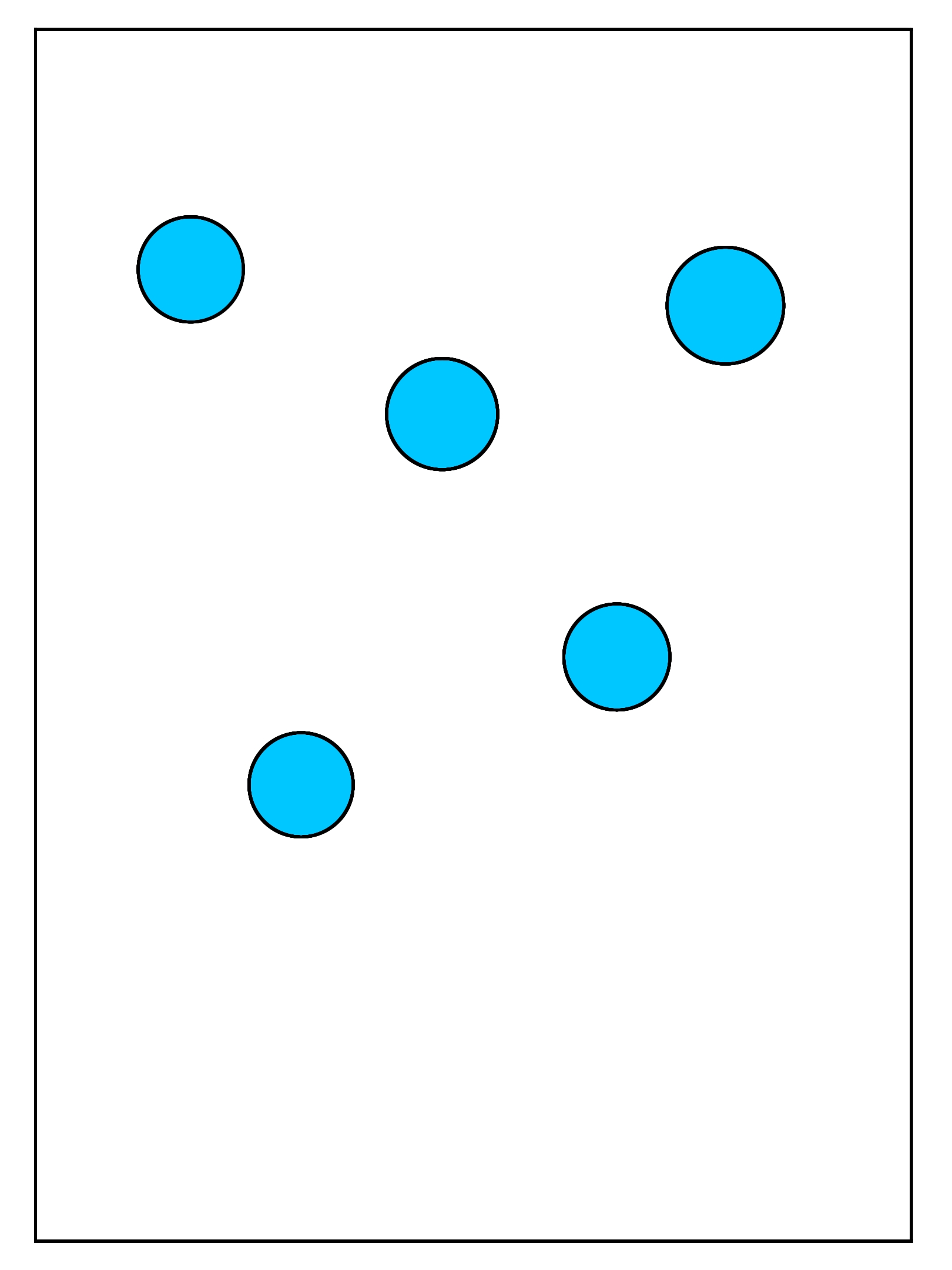}
        \subcaption{Zero-model}
        \label{fig: figure_zeromodel}
    \end{minipage}
    \begin{minipage}{.43\linewidth}
        \centering
        \includegraphics[width=0.5\linewidth]{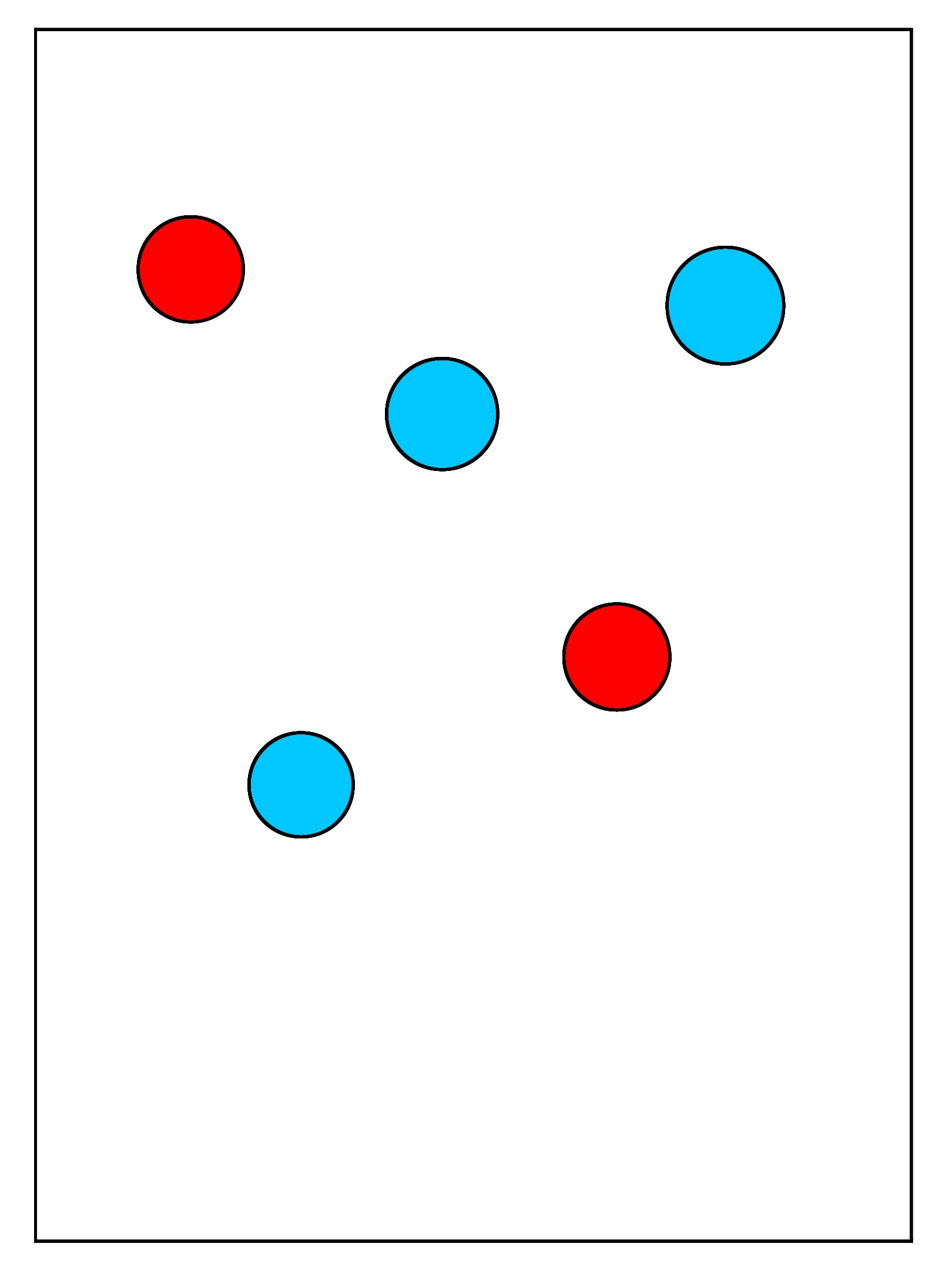}
        \subcaption{Non-zero-model}
        \label{fig: figure_nonzeromodel}
    \end{minipage}
    \caption{(a) is a zero-model of~(\ref{ex: ESQ}) and~(\ref{ex: DIS}), and (b) is a non-zero-model of~(\ref{ex: ESQ}) and~(\ref{ex: DIS}).}
    \label{fig: example of zero-model and non-zero-model}
\end{figure}

Crucially, the conclusions of these types of inferences become valid once we ignore the situations where the premise is vacuously true due to an empty set: namely, the situations with no red circles for (\ref{ex: ESQ_pre}), and the situations where all circles are red/blue for (\ref{ex: DIS_pre}).
These ``empty'' configurations are referred to as \textit{zero-models}, and the human cognitive tendency to systematically ignore them is called the \textit{neglect-zero effect}~\citep{Aloni-2022}.
Graphical examples of (non-)zero-models for (\ref{ex: ESQ}) and (\ref{ex: DIS}) are given in Figure~\ref{fig: example of zero-model and non-zero-model}.
\citet{Aloni-2022} hypothesizes that this bias arises because humans tend to prefer concrete representations to abstract ones, which are generally more cognitively demanding, so as to reduce cognitive load during sentence processing.

Given this motivation, the primary question we address in this study is whether LLMs process ESQ and DIS through a common mechanism.
In addition, we aim to confirm that the common mechanism, if it exists, indeed corresponds to the neglect-zero effect, by investigating whether it is distinct from the mechanism underlying an inference unrelated to the neglect-zero effect.
Here, as an instance of such an inference, we consider \textit{upper-bounded scalar implicature} (UPP), which is exemplified below.

\begin{exe}
    \ex\label{ex: UPP} UPP
    \begin{xlist}
        \ex\label{ex: UPP_pre} Some of the circles are red.
        \ex\label{ex: UPP_ante} $\leadsto$ Not all of the circles are red.
    \end{xlist}
\end{exe}

\noindent
Although UPP is not derived from literal meanings alone, like ESQ and DIS, it differs from the two because it is driven by a mechanism distinct from the neglect-zero effect.
More specifically, UPP is an instance of \textit{scalar implicature}, where scalar expressions such as ``some'' and numerals induce an inference excluding potential alternatives that belong to a certain predefined scale~\citep{horn1984scalarimplicature}.
In (\ref{ex: UPP}), we can assume a scale $\langle$some, all$\rangle$ because a simple declarative sentence with ``all'' entails its counterpart with ``some.''
That is, ``all'' can be considered stronger than ``some.''
Here, given that ``some'' is used in (\ref{ex: UPP_pre}), the hearer can infer that the speaker does not intend to convey the stronger expression ``all,'' because they should have used it if all the circles were indeed red.
The upshot is that UPP results from the hearer's inference about the speaker's intentions, unlike ESQ and DIS.
Hence, by comparing UPP and ESQ/DIS, we can distinguish whether the observed behavior is due to a mechanism specific to the neglect-zero effect or due to a general mechanism for dealing with non-literal meaning.

\subsection{Structural Priming}\label{sec: priming}

\textit{Structural priming} is a human cognitive tendency to process a sentence in the same way as its preceding sentence if they are similar in terms of their semantic or syntactic structures~\citep{bock1986priming, pickering2008structuralpriming}.
The sentence processed first is called the \textit{prime}, and the subsequent sentence is called the \textit{target}.

Structural priming is considered to result from the human tendency to reuse a mechanism during sentence processing.
From this perspective, structural priming has been employed as indirect evidence for the existence of a common mechanism underlying different linguistic inferences.
For example,~\citet{BOTT2016117} have demonstrated that structural priming is observed in humans when sentences that induce scalar implicature are presented in both the prime and the target.
On the other hand, structural priming was not observed between inferences hypothesized to be explained by the neglect-zero effect and those explained by scalar implicature~\citep{MEYER2021104206}.

Following the same line of reasoning, if the underlying principle of ESQ and DIS is the neglect-zero effect, while that of UPP is scalar implicature, it follows that structural priming should occur between ESQ and DIS, and should not occur between the ESQ/DIS group and UPP.
Based on this idea,~\citet{klochowicz2025neglect} hypothesized that if the prime suppresses the neglect-zero effect, specifically by forcing the subject to consider a zero-model of ESQ or DIS, the subject will be more likely to consider the zero-model of ESQ in the target as well.
We illustrate this point further using examples~(\ref{ex: ESQ}) and~(\ref{ex: DIS}), repeated below.

\begin{exe}
    \exi{(\ref{ex: ESQ})}
    \begin{xlist}
        \ex \{Fewer than three / At most two\} circles are red.
        \ex $\leadsto$ There are some red circles.
    \end{xlist}
    \exi{(\ref{ex: DIS})}
    \begin{xlist}
        \ex Each circle is red or blue.
        \ex $\leadsto$ There is a red circle and a blue circle.
    \end{xlist}
\end{exe}

\noindent
Let us consider a simple task in which the subject judges whether sentences like~(\ref{ex: ESQ_pre}) or~(\ref{ex: DIS_pre}) describe a given situation appropriately.
First, we use a stimulus where the situation is a zero-model and the prime sentence is~(\ref{ex: ESQ_pre}).
If the subject judges the description in the prime as appropriate, the cognitive act of considering the zero-model occurs.
Selecting the zero-model as the answer here suppresses the neglect-zero effect in the subsequent target.
Therefore, when the subject processes the target stimulus, provided that the target sentence is subject to inferences arising from the neglect-zero effect, we expect that the effect of considering the zero-model (i.e., suppression of the neglect-zero effect) in the prime continues.
Consequently, if the target situation corresponds to a zero-model, we can expect a higher probability of the model judging it as true.

Although the above procedure is designed for humans, we believe that it is also applicable to LLMs.
For instance,~\citet{jumelet-etal-2024-language} confirmed a syntactic structural priming effect on two dative constructions in LLMs, suggesting that LLMs are sensitive to structural priming.
Thus, we can expect that the priming-based experimental framework of \citet{klochowicz2025neglect}, together with an effective prompting strategy, will enable us to reveal whether LLMs exhibit the neglect-zero effect.

\section{Method}\label{chap: method}

\subsection{Task Setting}\label{sec: humanmethod}

Before describing the experimental method in detail, we provide an overview of the task setting of \citet{klochowicz2025neglect}.
The experiment involves a picture-matching task, where the subject performs two kinds of inferences, one in the prime and the other in the target.
The goal here is to test for the structural priming effect, thereby determining whether the two inferences are driven by a common mechanism.
More specifically, the prime involves one of the three inferences $I_p\in\left\{\text{ESQ},\text{DIS},\text{UPP}\right\}$, while the inference in the target is fixed to ESQ.
In a trial with the prime sentence, we ``suppress'' the $I_p$ inference by forcing the subject to choose a picture that contradicts the conclusion of $I_p$ but is true based on the literal meaning of the prime sentence.
Then, if ESQ is also suppressed in a trial with the target sentence, this result supports the conclusion that $I_p$ and ESQ are processed by a shared inference mechanism.
In this way, we can use structural priming to investigate whether any two of ESQ, DIS, and UPP share a common mechanism.

Having provided the general idea behind the experimental framework, we proceed to the details of how the experimental material is constructed.

\subsubsection{The Better-Picture Paradigm}\label{subsec: betterpicture}

The paradigm of this experiment, which we refer to as the \textit{better-picture} paradigm, involves a version of the picture-matching task.
It is a two-alternative forced-choice task with two pictures (named ``Picture 1'' and ``Picture 2'') and a single sentence.
A single execution of this task is referred to as a \textit{trial}, and the trials are divided into two types based on the type of pictures.
In the first type of trials, two open pictures and a sentence are presented, and the subject answers which picture matches the sentence (see Figure \ref{fig: example_of_open-picture_trial}).
In the second type of trials, an open picture and a covered picture are presented, and the covered picture is called the \textit{better-picture} (see Figure \ref{fig: example_of_better-picture_trial}).
When the subject addresses this type of trial, the better-picture has to be chosen if and only if the open picture does not match the sentence.

Next, we describe the content of an open picture.
The open pictures feature two types of simple shapes, selected from the following: triangles, squares, crosses, hearts, and circles.
One shape is in the upper half, and the other shape is in the lower half.
In addition, the shapes are colored, and those in one half have a homogeneous color, and those in the other half have mixed colors.
The colors used in this experiment are blue, brown, gray, orange, black, purple, green, pink, and yellow.

\subsubsection{The Experimental Framework}\label{subsec: framework}

\begin{figure}[t]
    \centering
    \begin{subfigure}[b]{1.0\linewidth}
        \centering
        \includegraphics[width=1.0\linewidth]{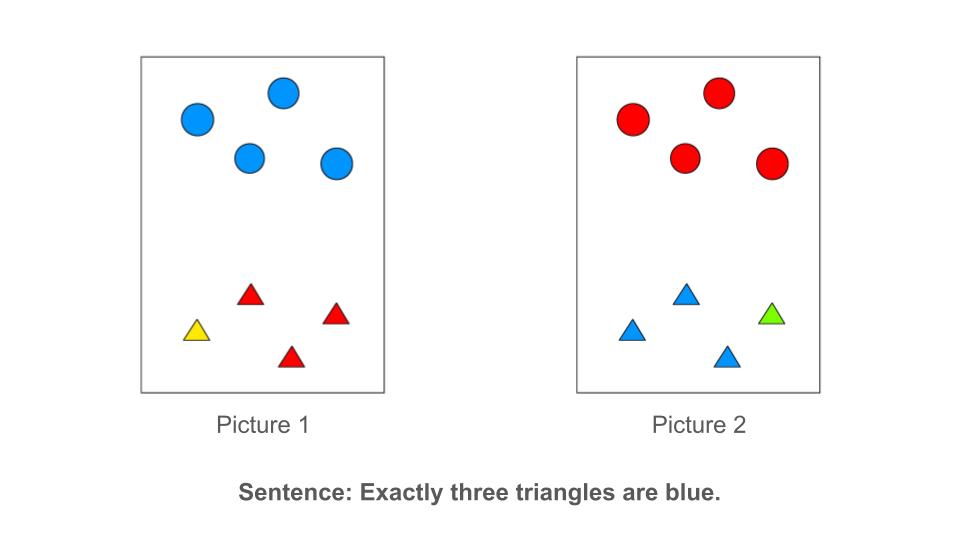}
        \caption{An example of a trial with two open pictures. The correct answer in this trial is Picture 2.}
        \label{fig: example_of_open-picture_trial}
        \vspace{10pt}
    \end{subfigure}
    \begin{subfigure}[b]{1.0\linewidth}
        \centering
        \includegraphics[width=1.0\linewidth]{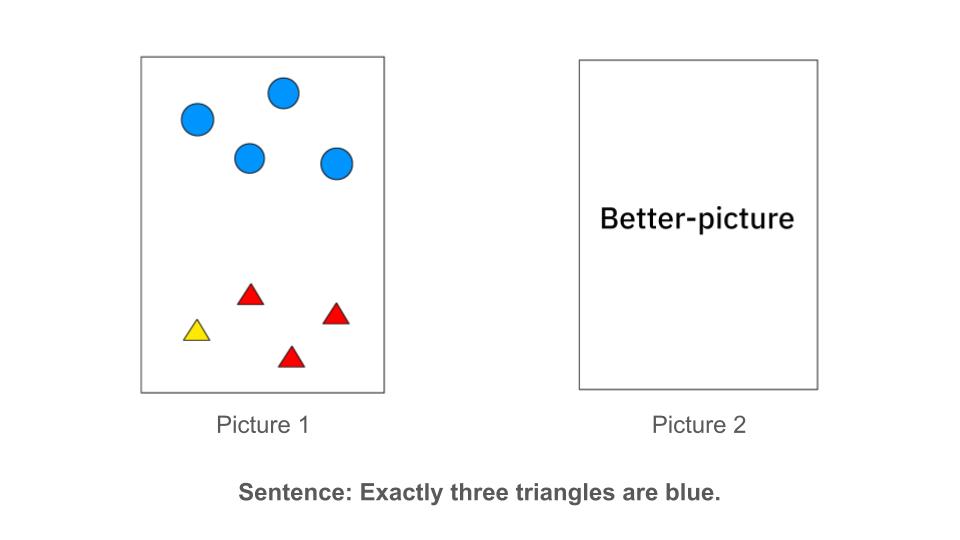}
        \caption{An example of a trial with a better-picture. The correct answer in this trial is Picture 2 (the better-picture), since Picture 1 is incompatible with the sentence.}
        \label{fig: example_of_better-picture_trial}
    \end{subfigure}
    \caption{Examples of trials.}
    \label{fig: example_of_open_and_better-picture_trials}
\end{figure}

In this experimental framework, there are three types of trials: \textit{prime trials}, \textit{target trials}, and \textit{filler trials}.
Prime trials and target trials correspond to the prime and the target, respectively (cf.\ Section~\ref{sec: priming}).
We group a sequence consisting of a prime trial and a target trial preceded by three filler trials as an \textit{experimental item}, where the filler trials separate different experimental items, ensuring that the target trial in the former experimental item does not induce structural priming for the prime trial in the latter experimental item.
In what follows, we explain the details of these three kinds of trials.

\begin{figure}[t]
    \centering
    \begin{subfigure}[b]{1.0\linewidth}
        \centering
        \includegraphics[width=1.0\linewidth]{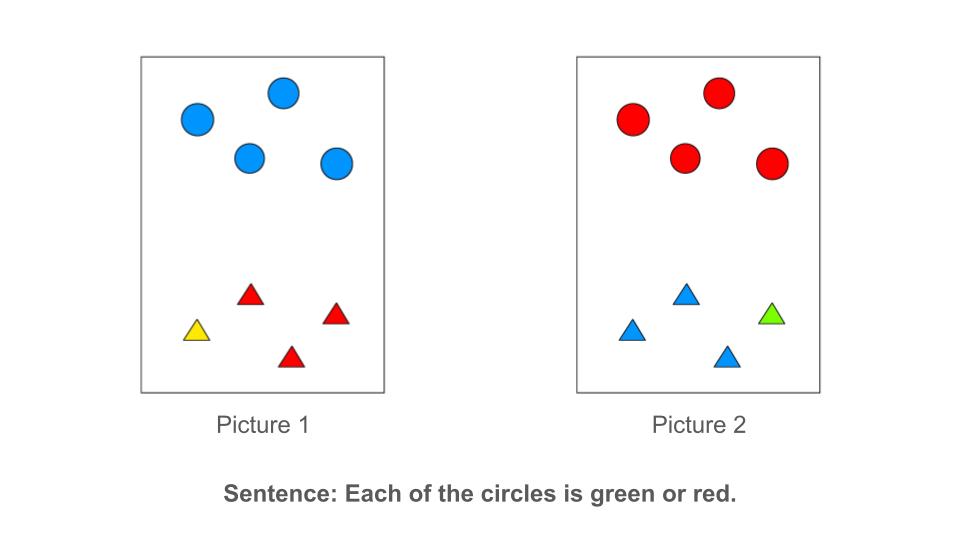}
        \caption{
            An example of a critical-prime trial.
            The correct answer in this trial is Picture 2 (a zero-model).
            Note that Picture 1 is designed to be incompatible with the sentence.
        }
        \label{fig: example_of_critical-prime_trial}
        \vspace{10pt}
    \end{subfigure}
    \begin{subfigure}[b]{1.0\linewidth}
        \centering
        \includegraphics[width=1.0\linewidth]{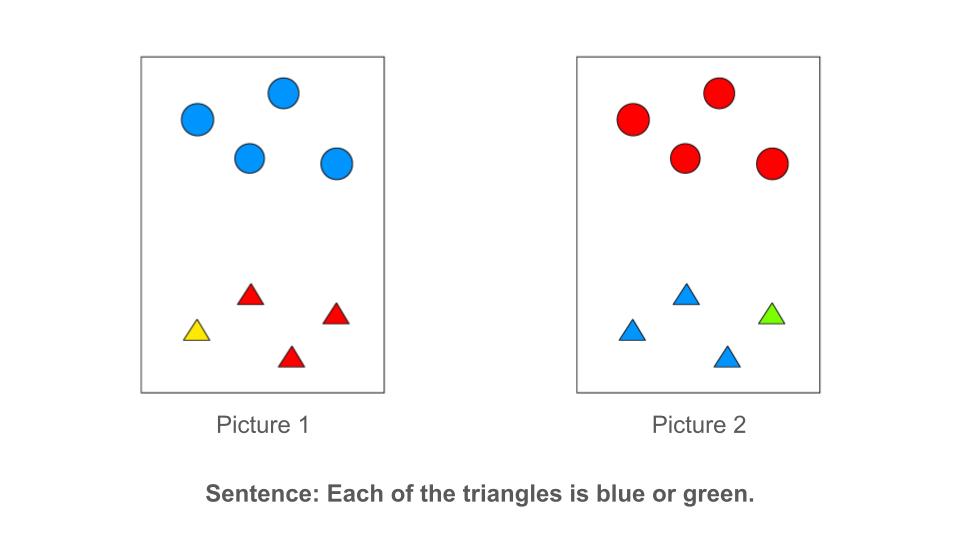}
        \caption{
            An example of a control-prime trial.
            The correct answer in this trial is Picture 2 (a non-zero-model).
        }
        \label{fig: example_of_control-prime_trial}
    \end{subfigure}
    \caption{Examples of a critical-prime trial and a control-prime trial ($I_p=\text{DIS}$).}
    \label{fig: example_of_critical_and_control-prime_trials}
\end{figure}

Prime trials, in which two open pictures are shown, are divided into two types based on whether the inference of interest is suppressed.
One is the \textit{critical-prime trial}, where Picture 1 does not match the sentence.
Here, the subject is expected to choose Picture 2, which shows a zero-model, that is, the situation neglected in the inference $I_p$ (see Figure~\ref{fig: example_of_critical-prime_trial}).
The other is the \textit{control-prime trial}, where Picture 1 does not match the sentence.
Again, the subject is expected to choose Picture 2, which shows a non-zero-model of the sentence (see Figure~\ref{fig: example_of_control-prime_trial}).
In the prime trials, Picture 1 is designed to be clearly incompatible with the sentence.
Since the task is a two-alternative forced-choice, this design ensures that Picture 2 is inevitably selected.
By comparing accuracy in target trials after critical-prime trials with that after control-prime trials, we can determine whether structural priming occurs.

\begin{figure}[t]
    \centering
    \includegraphics[width=1.0\linewidth]{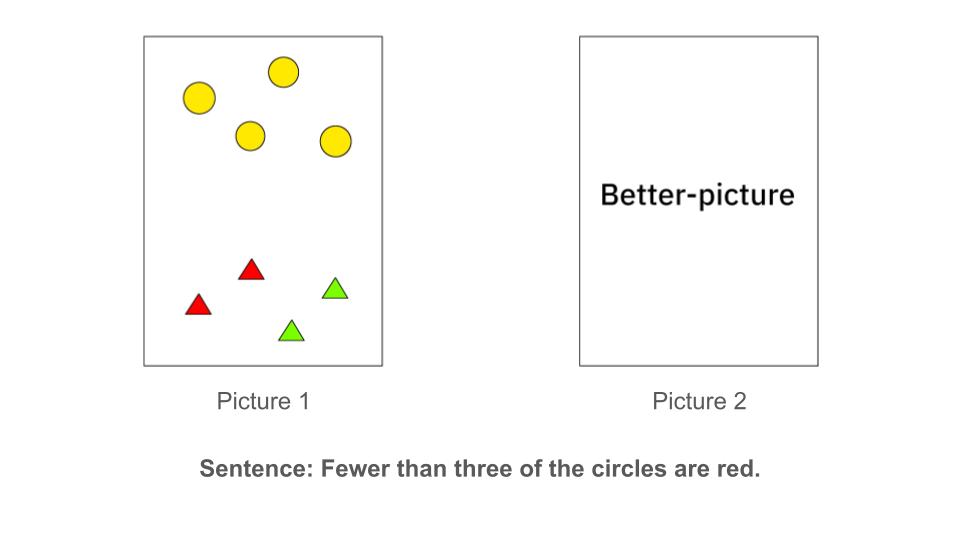}
    \caption{An example of a target trial. If priming occurs, Picture 1 should be selected, and if priming does not occur, Picture 2 should be selected.}
    \label{fig: example_of_target_trial}
\end{figure}

In target trials, one of the pictures is a better-picture, and the sentence includes the phrase ``fewer than three,'' which induces ESQ (see Figure~\ref {fig: example_of_target_trial}).
As the open picture shows a zero-model, it is expected that if structural priming is effective, the open picture will be chosen more often after the critical-prime trials than after the control-prime trials.

Finally, filler trials are designed so that the sentences in these trials do not induce the conversational implicatures of interest in this experiment (see Figure~\ref{fig: example_of_open_and_better-picture_trials} for an example of this trial).
In half of the prepared filler trials, Pictures 1 and 2 are open pictures, and the other half contain a better-picture.
In addition, the correct answer in trials containing a better-picture is either an open picture or a better-picture, and the number of each case is the same.

Here are further details of the experimental setting.
To prevent participants from predicting correct answers through memorization, the shapes and colors presented in the sentences and pictures are randomized for each trial.
Furthermore, to counterbalance the potential effects arising from features such as the color and shape of the pictures, four variations of the pair of a prime trial and a target trial are prepared, and one of them is selected almost randomly.\footnote{The choice is constrained to ensure an equal number of control-prime and critical-prime trials within a single sub-experiment.}
These variations are classified based on two criteria: whether they correspond to a zero-model and whether they are pictures that match the sentence, and as a result, we obtain the following four variations: $\langle$zero-model, answer$\rangle$, $\langle$zero-model, non-answer$\rangle$, $\langle$non-zero-model, answer$\rangle$, $\langle$non-zero-model, non-answer$\rangle$.

In addition, before the experimental items, a short training consisting of eight filler trials is conducted to help the subject get accustomed to the task.
After this training, 48 experimental items are presented to the subject.

\subsubsection{Four Types of Sub-Experiments}\label{subsec: subexperiment}

To consider how differently structural priming occurs among the three inferences,~\citet{klochowicz2025neglect} prepare four types of \textit{sub-experiments}: \textit{ESQ-sub-experiment}, \textit{DIS-sub-experiment}, \textit{UPP-sub-experiment}, and \textit{BAS-sub-experiment}.
These four sub-experiments have different types of prime trials, and the inference of interest is the same across all target trials.
All graphical examples for this section are shown in Appendix~\ref{app: examples_of_task}.

In ESQ-sub-experiments, the inference $I_p$ is ESQ, and the sentence of the prime trial includes the superlative empty-set quantifier ``at most two.''
In DIS-sub-experiments, the inference $I_p$ is DIS, and the sentence of the prime trial includes ``each.''
In UPP-sub-experiments, the inference $I_p$ is UPP, and the sentence of the prime trial includes the scalar ``some.''
As explained in Section~\ref{sec: neglect-zero}, UPP is an inference triggered by scalar expressions and considered to be based on a principle distinct from the neglect-zero effect.
Consequently, we expect this sub-experiment to yield results different from those observed in the ESQ and DIS sub-experiments.

In BAS-sub-experiments, the sentence of the prime trial is replaced by a filler trial, so it does not induce the inferences mentioned above.
Therefore, this sub-experiment serves as the baseline for this experiment in terms of the choice rate of zero-models in target trials.
In addition, by comparing this sub-experiment and the others, we can distinguish priming from general adaptation effects caused by repeated exposure to similar tasks in target trials.

\subsection{The Evaluation Protocol for LLMs}\label{sec: llmmethod}

Our evaluation protocol for LLMs follows the methodology of~\citet{tsvilodub2024experimental}, who have applied complex experimental designs for humans to LLMs.
Specifically, we convert the visual contents of the pictures into natural language text, which we then combine with a short instruction and a sentence to form a single trial.
In addition, we randomize the order of the answer options (``First'' and ``Second'').

Each sub-experiment is constructed by first presenting the task description as an instruction, followed by eight filler trials introduced as few-shot training, and finally arranging the trials as described in Section~\ref{subsec: framework}.

The prompt formats and examples are provided in Appendix~\ref{app: prompts}.
This conversion is automatically performed by a Python program, utilizing the data provided by~\citet{klochowicz2025neglect}.

\section{Experiments and Analysis}\label{chap: experiment}

\subsection{Settings}\label{sec: settings}

\subsubsection{Models and Prompts}\label{subsec: models}

We evaluate five open LLMs: four instruction-tuned models from the Gemma-3 series (1B, 4B, 12B, and 27B models)\footnote{\url{https://huggingface.co/collections/google/gemma-3-release}} and Llama-4-Scout-17B-16E-Instruct.\footnote{\url{https://huggingface.co/meta-llama/Llama-4-Scout-17B-16E-Instruct}}
Hereafter, we refer to Gemma-3-$n$B-it as ``Gemma-3-$n$B'' and Llama-4-Scout-17B-16E-Instruct as ``Llama-4.''
In addition, we evaluate one closed LLM, GPT-5 nano.\footnote{\url{https://developers.openai.com/api/docs/models/gpt-5-nano}}

We treat each seed value assigned to the LLMs as an individual participant in human experiments~\citep{mccurdy-etal-2020-inflecting}.
The total number of seeds is 80 per sub-experiment, for a total of 320 per model.

As mentioned in Section~\ref{sec: llmmethod}, we construct a dataset for our experiment with LLMs by using the method of~\citet{tsvilodub2024experimental} for converting images into texts.
The original dataset is provided by~\citet{klochowicz2025neglect}, and it contains 380 items for prime trials and target trials per sub-experiment, along with 250 items for filler trials.
To ensure that the items presented to each seed are not identical, prompts are constructed by randomly sampling from this dataset.

For each model, 80 seeds are prepared for each of the four sub-experiments.
The prompt for each seed consists of a few-shot training (consisting of eight filler trials) and 48 experimental items.
For specific examples of the prompts and the detailed construction methodology, see Appendix~\ref{app: prompts}.

\subsubsection{Metrics}\label{subsec: metrics}

\begin{table*}[tbp]
    \centering
    \small
    \begin{tabular}{cllllllllll}
        \toprule
         & \multicolumn{5}{c}{\textbf{Gemma-3-27B}} & \multicolumn{5}{c}{\textbf{Llama-4}}\\
        \cmidrule(lr){2-6} \cmidrule(lr){7-11}
        & Filler & \multicolumn{3}{c}{Prime} & Target & Filler & \multicolumn{3}{c}{Prime} & Target \\
        \cmidrule(lr){3-5} \cmidrule(lr){8-10}
        &  & Critical & Control & Overall &  &  & Critical & Control & Overall &  \\
        \midrule
        ESQ & 74.9 & 99.9 & 95.4 & 97.6 & 97.1 & 91.3 & 96.0 & 99.4 & 97.7 & 74.1 \\
        DIS & 71.1 & 98.2 & 97.5 & 97.8 & 95.6 & 91.6 & 92.3 & 95.6 & 94.0 & 72.3 \\
        UPP & 72.8 & 100.0 & 91.0 & 95.5 & 96.2 & 89.8 & 98.3 & 84.4 & 91.6 & 29.8 \\
        BAS & 76.0 & - & - & 93.8 & 97.2 & 90.7 & - & - & 97.2 & 49.5 \\
        \bottomrule
    \end{tabular}

    \vspace{1em}

    \begin{tabular}{cllllllllll}
        \toprule
         & \multicolumn{5}{c}{\textbf{GPT-5 nano}}
         & \multicolumn{5}{c}{\textbf{Humans}}\\
        \cmidrule(lr){2-6} \cmidrule(lr){7-11}
        & Filler & \multicolumn{3}{c}{Prime} & Target & Filler & \multicolumn{3}{c}{Prime} & Target\\
        \cmidrule(lr){3-5} \cmidrule(lr){8-10}
        &  & Critical & Control & Overall &  &  & Critical & Control & Overall &  \\
        \midrule
        ESQ & 96.5 & 99.9 & 53.4 & 77.1 & 98.8 & 84.0 & 83.2 & 85.6 & 84.4 & 64.7 \\
        DIS & 96.8 & 98.1 & 97.6 & 97.9 & 98.5 & 85.0 & 98.7 & 95.8 & 97.2 & 44.0 \\
        UPP & 96.8 & 98.7 & 92.1 & 95.4 & 99.0 & 85.3 & 99.0 & 97.4 & 98.1 & 38.7 \\
        BAS & 96.8 & - & - & 99.9 & 99.1 & 85.7 & - & - & 97.6 & 49.0 \\
        \bottomrule
    \end{tabular}
    \caption{The mean accuracy in filler trials and the rate of choosing a zero-model in prime and target trials in each sub-experiment of Gemma-3-27B, Llama-4, GPT-5 nano, and humans. Human data were sourced from~\citet{klochowicz2025neglect}.}
    \label{table: accuracy_result}
\end{table*}

In this experiment, we use the accuracy of LLMs' responses as a metric.
Following~\citet{tsvilodub2024experimental}, the seed values that achieve an accuracy below 75\% in filler trials are excluded.
We also exclude experimental items if the answer in their prime trial is incorrect.
We refer to the target trials that survive these exclusion steps as \textit{valid target trials}.

For each sub-experiment, we calculate the accuracy of valid target trials under three conditions: following the critical-prime trials, following the control-prime trials, and following all prime trials.
In the BAS-sub-experiment, as there is no distinction between control-prime and critical-prime trials, we only calculate accuracy following all prime trials.
Based on these data, we conduct a statistical analysis using a \textit{generalized linear mixed model} (GLMM) for each sub-experiment.
See Appendix~\ref{app: glmm} for a detailed description of the GLMM.

In this experiment, a GLMM is expressed by the following equation:
\begin{align*}
    \text{logit}(q_i) = \beta_i + \sum_{j}{\beta_{ji}x_{ji}} + \sum_k{r_{ki}}.
\end{align*}
Here, $q_i$ is the mean of the response variable, $\text{logit}(\cdot)$ is the logit function, $\beta_i$ and $\beta_{ji}$ are the parameters of this model, $x_{ji}$ are the fixed effects, and $r_j$ are the random effects.
In our experiment, the response variable corresponds to the rate of choosing a zero-model in valid target trials.
The fixed effects correspond to the effect of the critical-prime trials relative to the control-prime trials, which we call the \textit{priming condition}, and the number indicating the order of the experimental items, which we call the \textit{trial index}.
The random effects correspond to the intercepts and slopes for each seed, as well as the intercepts for each experimental item.

Through this analysis, we investigate how each fixed effect influences the accuracy in each sub-experiment.
Furthermore, by including the effect of each sub-experiment relative to the BAS-sub-experiment as fixed effects, we examine whether there are differences in accuracy between the BAS-sub-experiment and the other sub-experiments.
For this comparison, we use the effect of the control-prime trials in each sub-experiment.

\subsection{Results and Analysis}\label{sec: results and analysis}

Table~\ref{table: accuracy_result} summarizes the mean accuracy across seeds for filler trials in each sub-experiment of Gemma-3-27B, Llama-4, GPT-5 nano, and humans, along with the rate of choosing a zero-model in prime trials (critical-prime only, control-prime only, and both) and target trials.
The filler accuracy is calculated from data from all seeds, and all other values are calculated solely from the seeds retained after exclusion based on filler accuracy.
Note that all values are rounded to one decimal place.

Regarding Gemma-3-1B, Gemma-3-4B, and Gemma-3-12B, all seeds were excluded because their accuracy on filler trials was under 75\% across all seeds.
Their results are shown in Appendix~\ref{app: human_results}.
Regarding Gemma-3-27B, 192 seeds were excluded, consisting of 45 from the ESQ-sub-experiment, 57 from the DIS-sub-experiment, 50 from the UPP-sub-experiment, and 40 from the BAS-sub-experiment.
In contrast, we used all seeds for Llama-4 and GPT-5 nano.

Note that there were rare cases where the model generated statements such as ``Both, but I must choose one, \{First/Second\}.''
In such cases, the final choice mentioned was treated as the model's response.
Furthermore, any other trials that did not yield a clear answer were treated as failed.

\subsubsection{Statistical Analysis}\label{subsec: llama4_analysis}

In the following, we conduct a statistical analysis of the results for each model.

\begin{figure*}
    \centering
    \begin{subfigure}[b]{0.32\linewidth}
        \centering
        \includegraphics[width=1.0\linewidth]{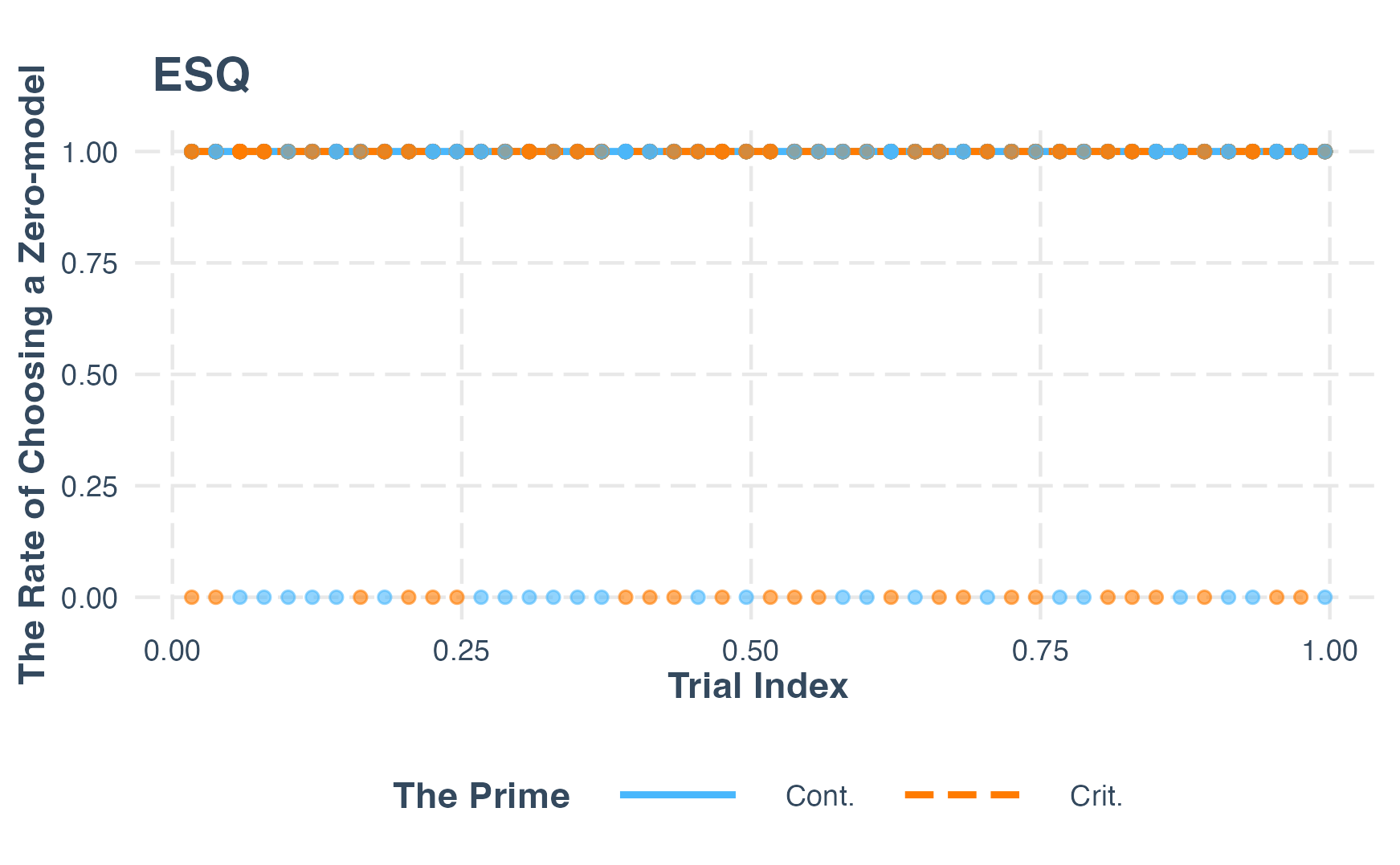}
        \caption{ESQ-sub-experiment}
        \label{fig: esq_gemma}
    \end{subfigure}
    \begin{subfigure}[b]{0.32\linewidth}
        \centering
        \includegraphics[width=1.0\linewidth]{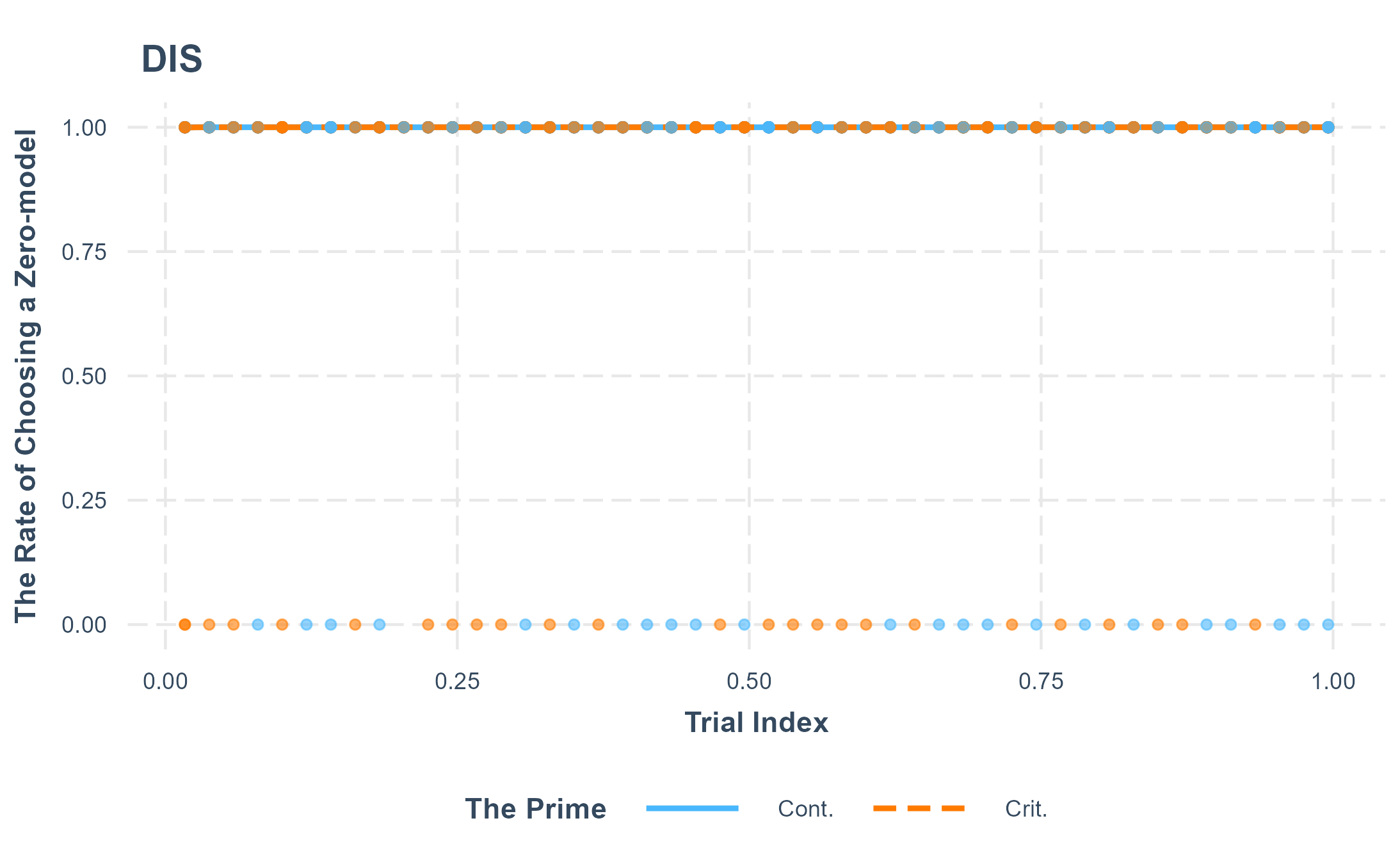}
        \caption{DIS-sub-experiment}
        \label{fig: dis_gemma}
    \end{subfigure}
    \begin{subfigure}[b]{0.32\linewidth}
        \centering
        \includegraphics[width=1.0\linewidth]{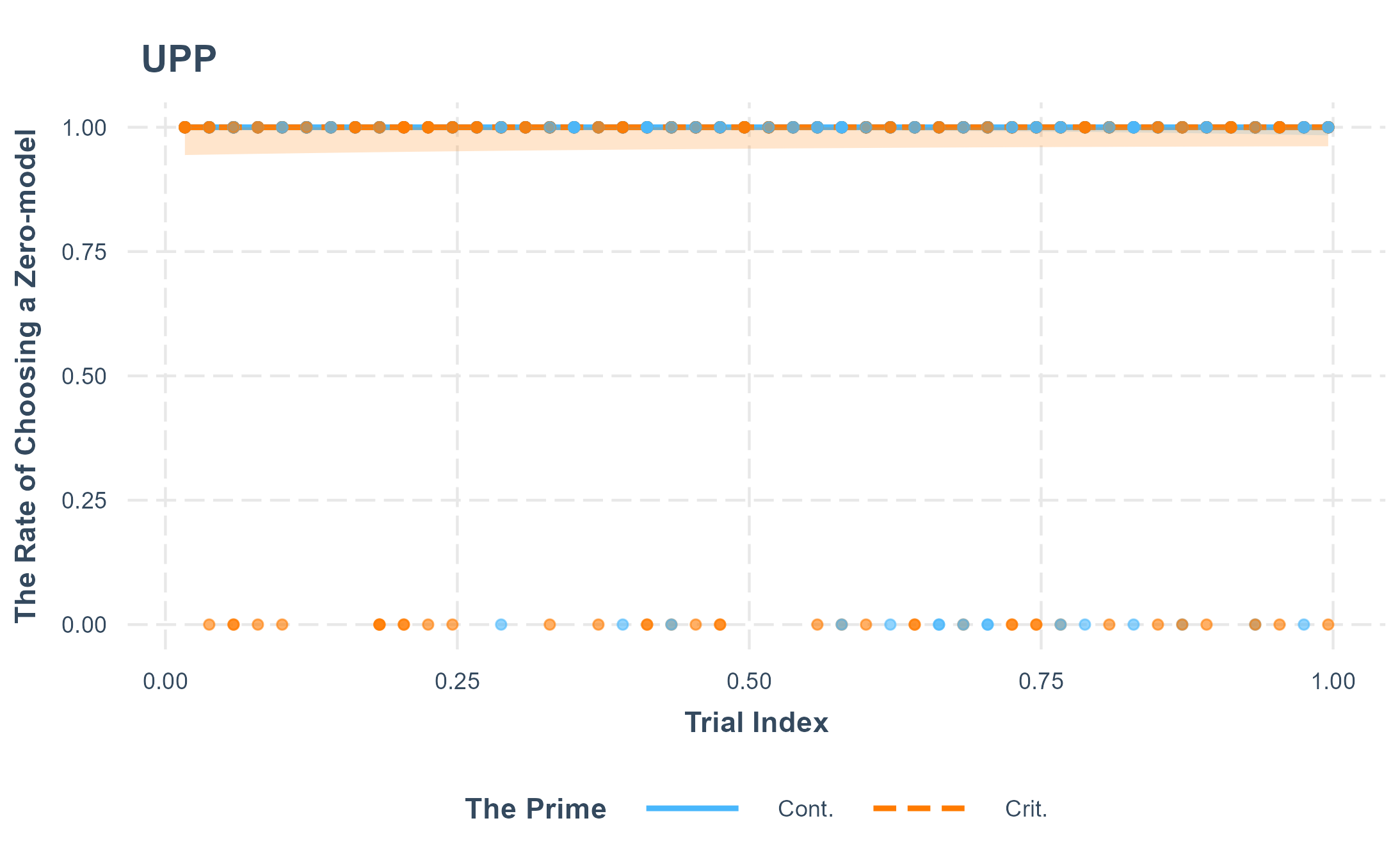}
        \caption{UPP-sub-experiment}
        \label{fig: upp_gemma}
    \end{subfigure}
    \caption{The regression curves for GLMM in each sub-experiment of Gemma-3-27B. It should be noted that we determined the results of this analysis to be inappropriate.}
    \label{fig: gemma3_glmm}
\end{figure*}

\begin{figure*}
    \centering
    \begin{subfigure}[b]{0.32\linewidth}
        \centering
        \includegraphics[width=1.0\linewidth]{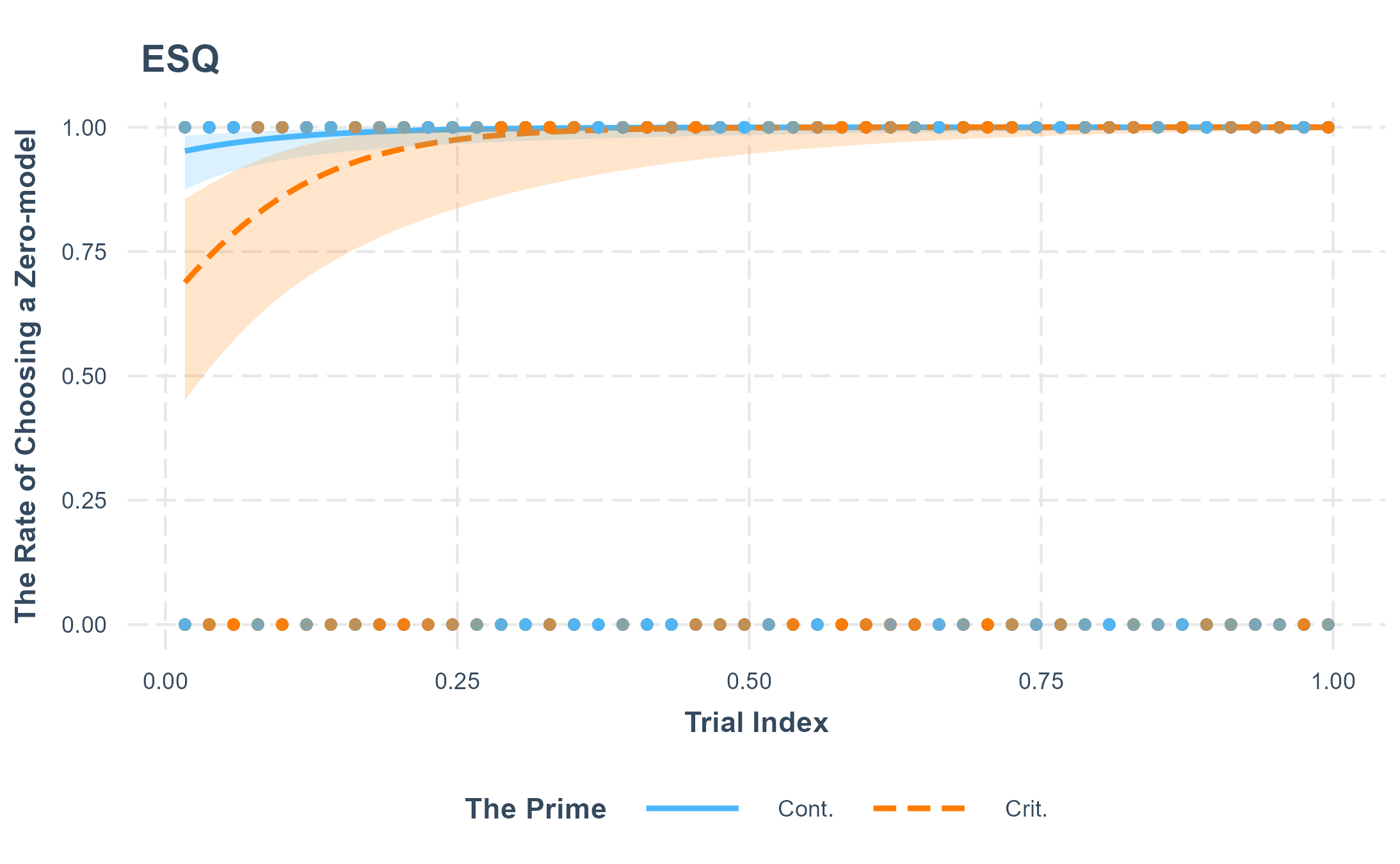}
        \caption{ESQ-sub-experiment}
        \label{fig: esq_llama}
    \end{subfigure}
    \begin{subfigure}[b]{0.32\linewidth}
        \centering
        \includegraphics[width=1.0\linewidth]{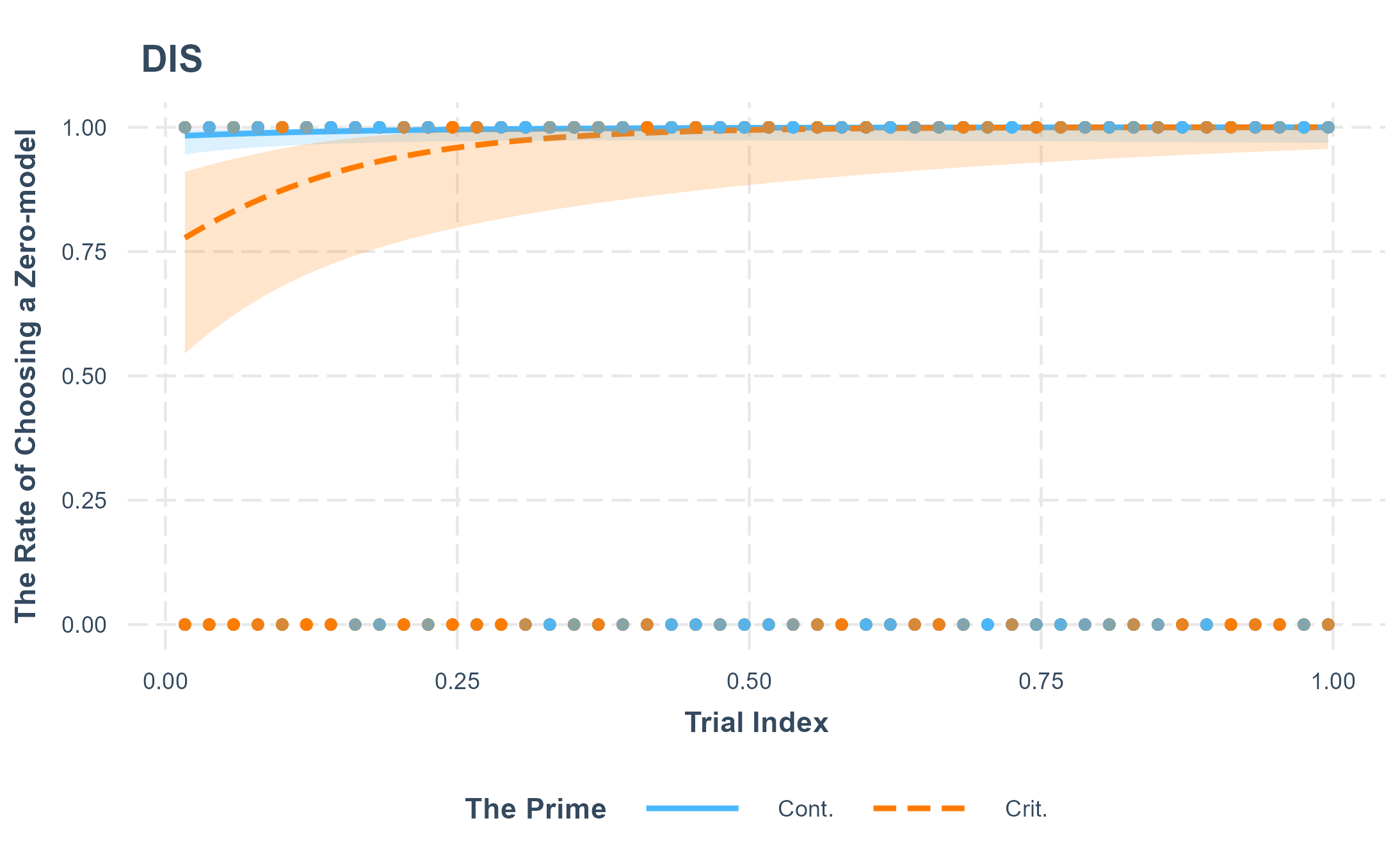}
        \caption{DIS-sub-experiment}
        \label{fig: dis_llama}
    \end{subfigure}
    \begin{subfigure}[b]{0.32\linewidth}
        \centering
        \includegraphics[width=1.0\linewidth]{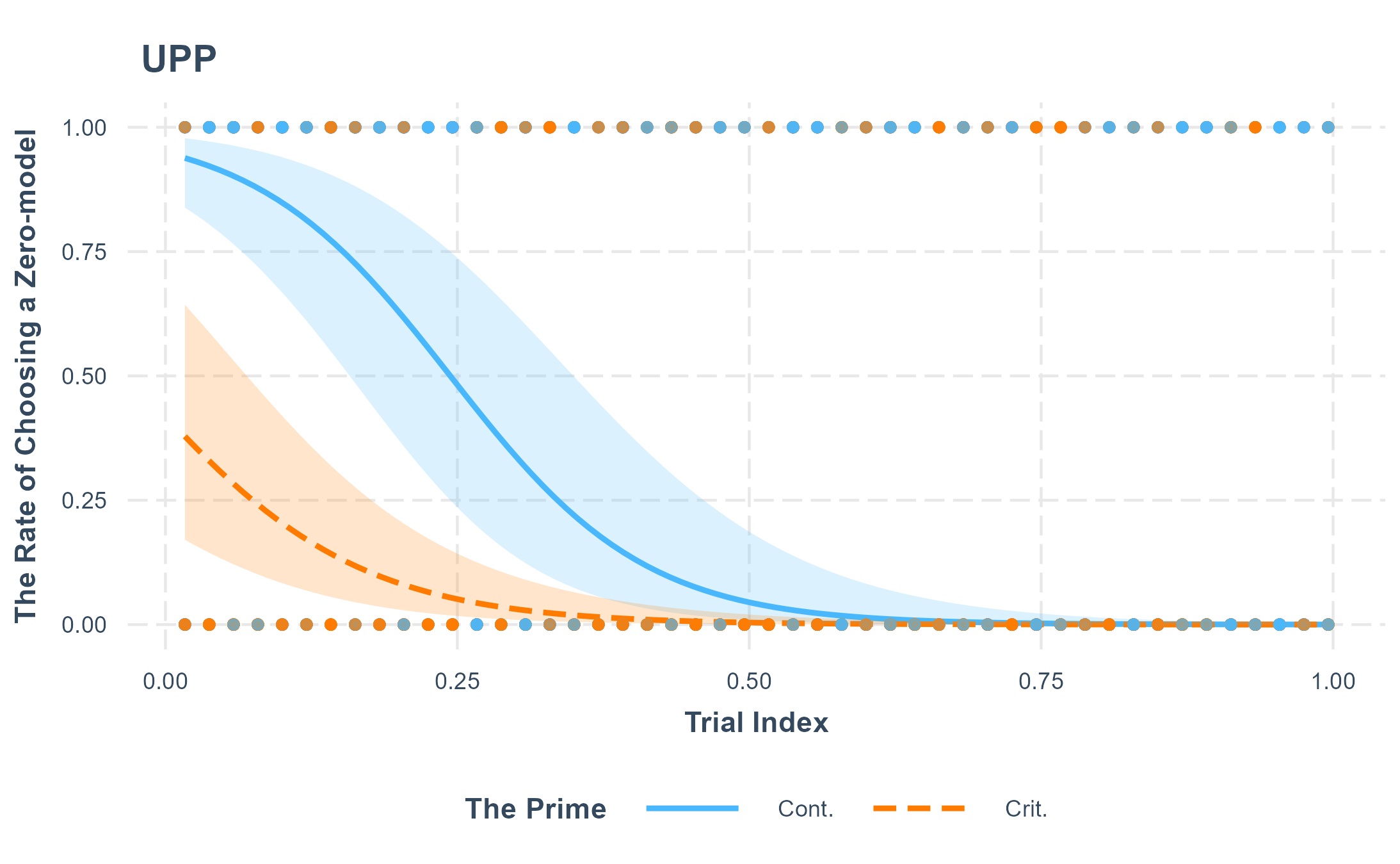}
        \caption{UPP-sub-experiment}
        \label{fig: upp_llama}
    \end{subfigure}
    \caption{The regression curves for GLMM in each sub-experiment of Llama-4.}
    \label{fig: llama4_glmm}
\end{figure*}

\begin{table}[tbp]
  \centering
  \small
    \begin{tabular}{lllll}
      \toprule
       & BAS & Index & Prime & Index \& Prime \\
      \midrule
      ESQ & positive & positive & negative & positive \\
      DIS & ns & positive & negative & positive \\
      UPP & positive & ns & negative & ns \\
      BAS & - & ns & - & - \\
      \bottomrule
    \end{tabular}
    \caption{Summary of whether each fixed effect is significantly correlated with the observed data in the Gemma-3-27B experiment, and if so, whether the correlation is positive or negative. ``ns'' denotes non-significant.}
    \label{table: analysis_gemma3}
\end{table}

\paragraph*{Gemma-3-27B}
Figure~\ref{fig: gemma3_glmm} shows the regression curves derived from GLMM for each sub-experiment of Gemma-3-27B.
Table~\ref{table: analysis_gemma3} summarizes whether each fixed effect is significantly correlated with the observed data, and if so, whether the correlation is positive or negative.

First, a general characteristic observed in all sub-experiments is that priming had a significant negative effect.
That is, once it was forced to select a zero-model in the prime, Gemma-3-27B became less likely to select one in the target.
This result indicates that Gemma-3-27B does not exhibit the neglect-zero effect.

However, since Table~\ref{table: analysis_gemma3} shows that the model becomes significantly less likely to consider zero-models after initially taking them into account, it is evident that the model is sensitive to zero-models to some extent.
These results suggest that semantic structural priming of zero-models in Gemma-3-27B manifests in a manner different from that in humans, where zero-models become more likely to be chosen after considering zero-models.
This point needs further detailed investigation.

In addition, because the accuracy in target trials is very high, it is possible that Gemma-3-27B rarely performs inference based on non-literal meanings and primarily processes sentences based on their literal meanings.
It is worth noting that~\citet{yerukola-etal-2024-pope} reported that LLMs struggle to generate responses based on non-literal meanings, which is consistent with our result here.

\begin{table}[tbp]
  \centering
  \small
    \begin{tabular}{lllll}
      \toprule
       & BAS  & Index & Prime & Index \& Prime \\
      \midrule
      ESQ & positive & positive & negative & positive \\
      DIS & positive & ns & negative & positive \\
      UPP & positive & negative & negative & positive \\
      BAS & - & negative & - & - \\
      \bottomrule
    \end{tabular}
    \caption{Summary of whether each fixed effect is significantly correlated with the observed data in the Llama-4 experiment, and if so, whether the correlation is positive or negative.
    ``ns'' denotes non-significant.}
  \label{table: analysis_llama4}
\end{table}

\paragraph*{Llama-4}
Figure~\ref{fig: llama4_glmm} shows the regression curves derived from GLMM for each sub-experiment of Llama-4.
Table~\ref{table: analysis_llama4} summarizes whether each fixed effect is significantly correlated with the observed data, and if so, whether the correlation is positive or negative.

As with Gemma-3-27B, priming had a significant negative effect in all sub-experiments.
This result indicates that Llama-4 does not exhibit the neglect-zero effect.

In addition, as indicated by the results in Figure~\ref{fig: llama4_glmm}, the behavior observed in UPP is apparently different from that in ESQ and DIS.
This suggests that a different underlying mechanism is at play in UPP compared to ESQ and DIS.

Taken together, these findings imply that semantic structural priming in Llama-4 takes a form distinct from that in humans, and, furthermore, that the neglect-zero effect is unlikely to manifest.

\begin{figure*}
    \centering
    \begin{subfigure}[b]{0.32\linewidth}
        \centering
        \includegraphics[width=1.0\linewidth]{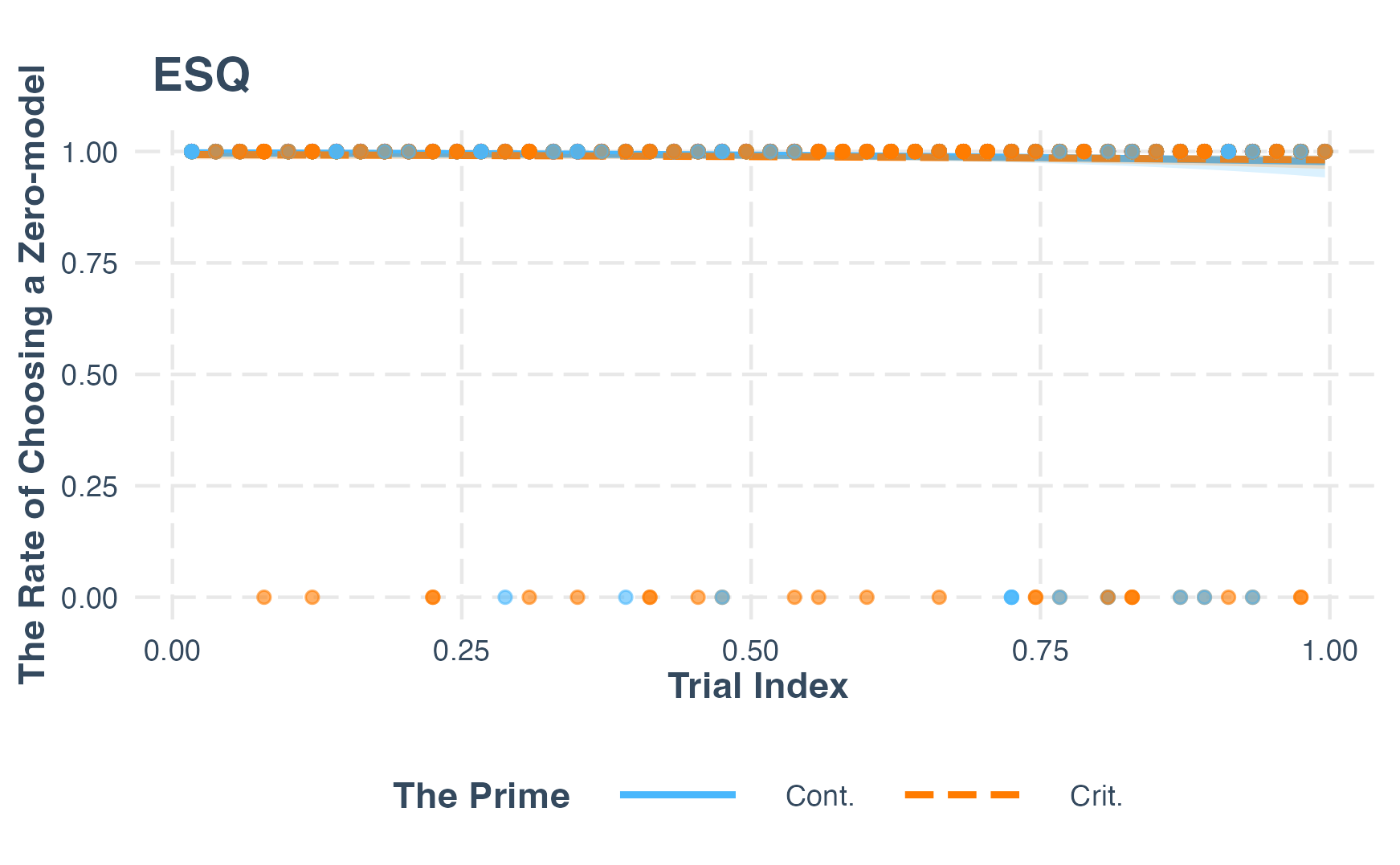}
        \caption{ESQ-sub-experiment}
        \label{fig: esq_gpt}
    \end{subfigure}
    \begin{subfigure}[b]{0.32\linewidth}
        \centering
        \includegraphics[width=1.0\linewidth]{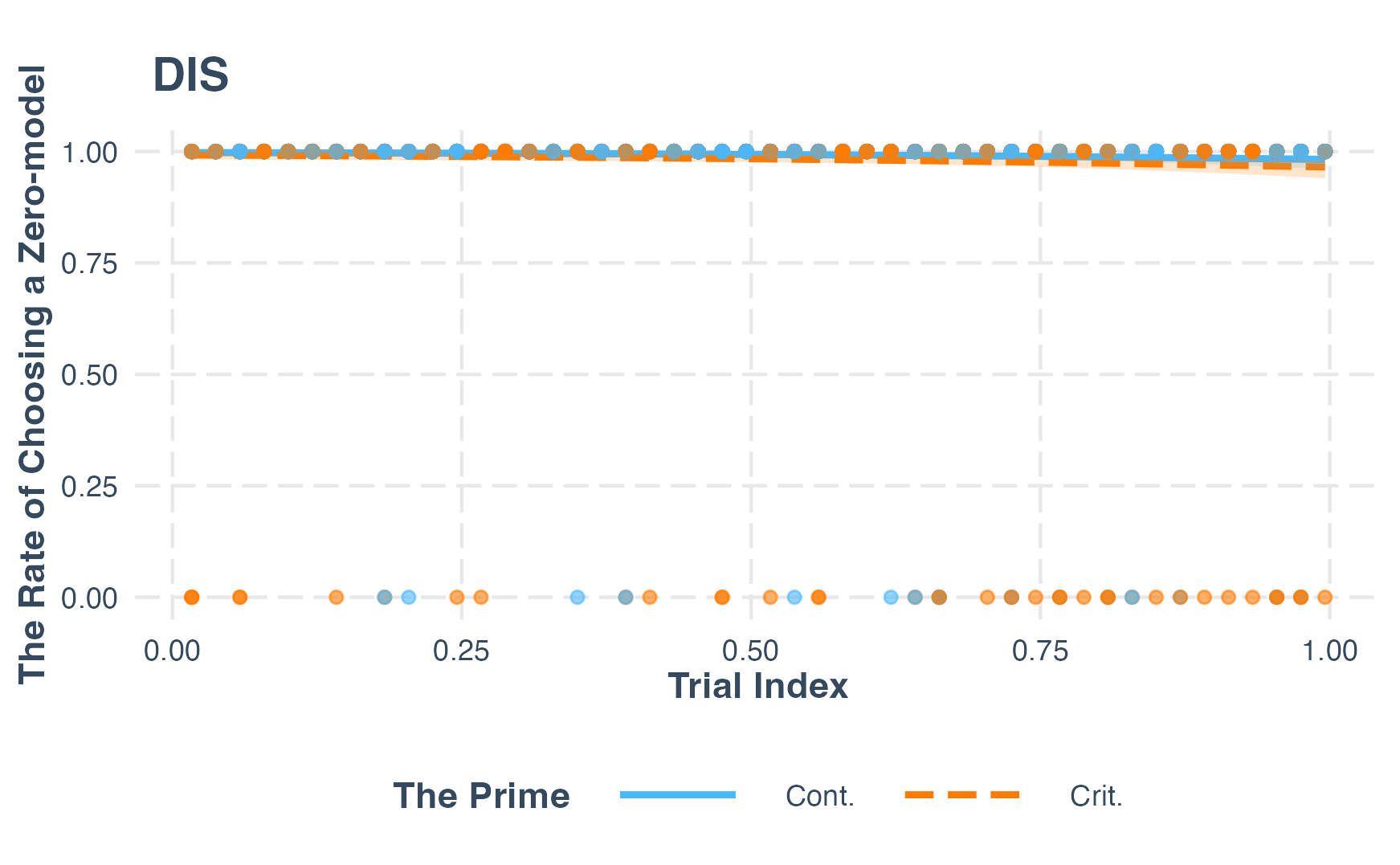}
        \caption{DIS-sub-experiment}
        \label{fig: dis_gpt}
    \end{subfigure}
    \begin{subfigure}[b]{0.32\linewidth}
        \centering
        \includegraphics[width=1.0\linewidth]{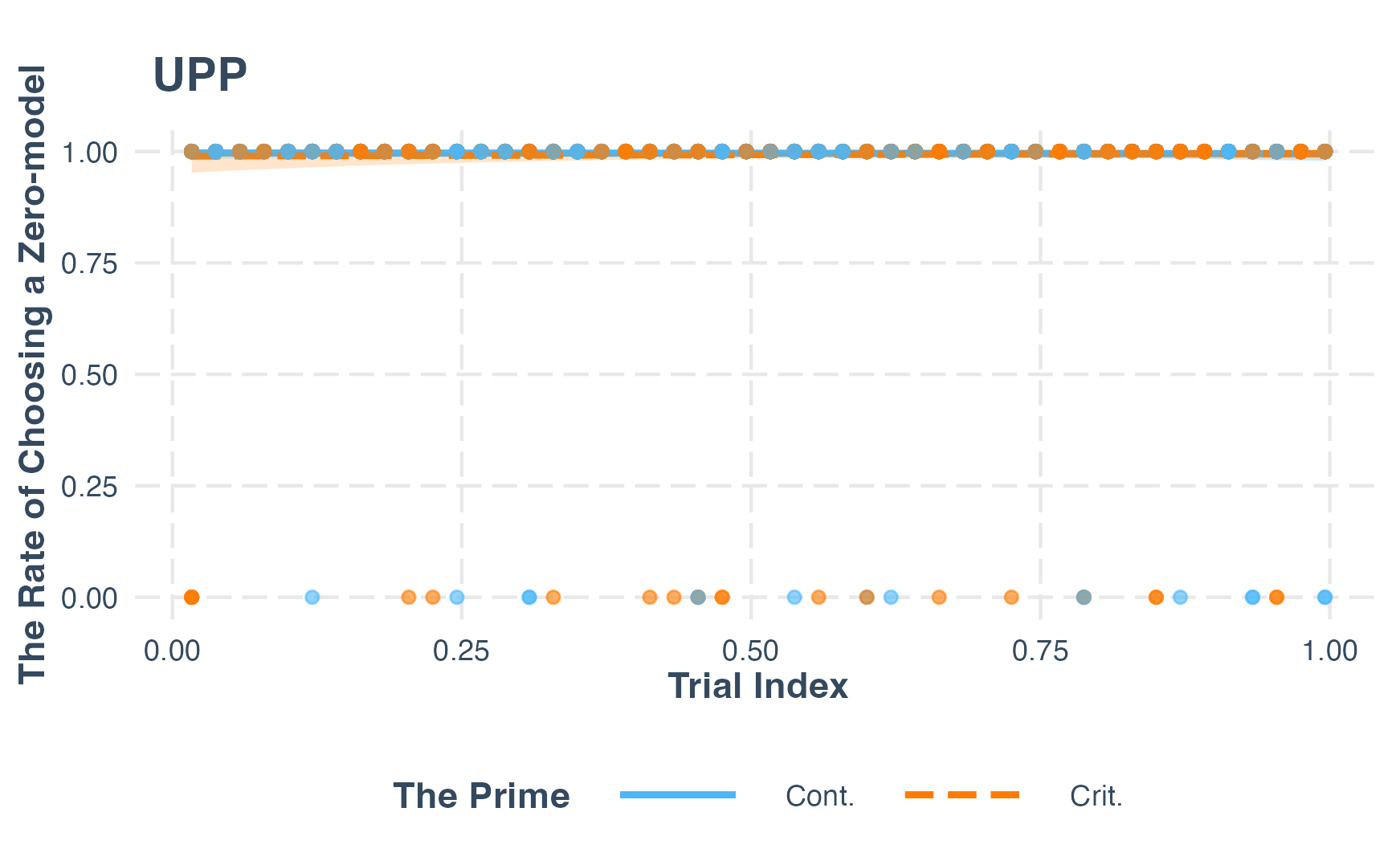}
        \caption{UPP-sub-experiment}
        \label{fig: upp_gpt}
    \end{subfigure}
    \caption{The regression curves for GLMM in each sub-experiment of GPT-5 nano.}
    \label{fig: gpt_glmm}
\end{figure*}

\paragraph*{GPT-5 nano}
Figure~\ref{fig: gpt_glmm} shows the regression curves derived from GLMM for each sub-experiment of GPT-5 nano.
Table~\ref{table: analysis_gpt} summarizes whether each fixed effect is significantly correlated with the observed data, and if so, whether the correlation is positive or negative.

First, priming had a non-significant effect in all sub-experiments.
This result indicates that GPT-5 nano does not exhibit the neglect-zero effect and is not sensitive to zero-models.

In addition, as with Gemma-3-27B, due to the high accuracy across all sub-experiments, this model may place a greater emphasis on literal meaning during its inference process.
On the other hand, as shown in Table~\ref{table: accuracy_result}, the accuracy rate in the control-prime trials is lower than that in the critical-prime trials, which is particularly prominent in ESQ.
Further investigation is necessary to elucidate the underlying causes of this phenomenon.

\begin{table}[tbp]
  \centering
  \small
    \begin{tabular}{lllll}
      \toprule
       & BAS  & Index & Prime & Index \& Prime \\
      \midrule
      ESQ & ns & ns & ns & ns \\
      DIS & ns & negative & ns & ns \\
      UPP & ns & ns & ns & ns \\
      BAS & - & ns & - & - \\
      \bottomrule
    \end{tabular}
    \caption{Summary of whether each fixed effect is significantly correlated with the observed data in the GPT-5 nano experiment, and if so, whether the correlation is positive or negative. ``ns'' denotes non-significant.}
  \label{table: analysis_gpt}
\end{table}

\section{Conclusion}\label{chap: conclusion}

In this study, against the backdrop of growing interest in the similarities between the cognitive processes of LLMs and those of humans, we focused on a specific human cognitive bias known as the neglect-zero effect.
To determine whether this effect is also observed in LLMs, we designed a framework for analyzing LLMs using structural priming by combining the methods of~\citet{klochowicz2025neglect} and~\citet{tsvilodub2024experimental}, and conducted experiments on six models utilizing this framework.

Our experimental results suggest that these models do not exhibit the neglect-zero effect in their sentence processing, unlike humans.
More specifically, the results from Gemma-3-27B and GPT-5 nano indicate that some models tend to primarily process the literal meaning instead of considering the non-literal meaning.
On the other hand, the results from Gemma-3-27B and Llama-4 show that some models may be sensitive to zero-models in a non-human-like manner.

Given these findings, an important direction for future work would be to conduct experiments in other settings.
For example, it would be valuable to conduct experiments on a larger number of models, or to use experimental settings capable of distinguishing between the manifestation of structural priming effects specific to LLMs and the disregard for non-literal meanings.
Such studies will reveal whether a unified explanation for the neglect-zero effect in LLMs is possible.

\section*{Limitations}

We observed a significant difference in the results between Gemma-3-27B/GPT-5 nano and Llama-4.
In this study, we attributed the results to the possibility that Gemma-3-27B and GPT-5 nano place greater weight on literal meaning than non-literal meaning.
However, to analyze this difference more precisely, it is necessary to conduct similar experiments on a larger number of models.

\section*{Acknowledgments}

This work was supported by JSPS KAKENHI Grant Number JP24H00809 and JST CREST Grant Number JPMJCR2565, Japan.

\bibliography{custom}

\appendix

\clearpage

\begin{table*}
  \centering
    \begin{tabular}{llll}
      \toprule
      Sub-experiment & Type of trial & Example sentence & Pictures \\
      \midrule
      ESQ & critical-prime & At most two of the circles are blue. & A \\
      ESQ & control-prime & At most two of the triangles are green. & A* \\
      DIS & critical-prime & Each of the circles is green or red. & A \\
      DIS & control-prime & Each of the triangles is blue or green. & A \\
      UPP & critical-prime & Some of the hearts are red. & A \\
      UPP & control-prime & Some of the triangles are blue. & A \\
      BAS & control-prime & Exactly three triangles are blue. & A \\
      All & target & Fewer than three of the circles are red. & B \\
      \bottomrule
    \end{tabular}
    \caption{Examples of experimental items in each sub-experiment. The alphabet in the Pictures column is based on Figure~\ref{fig: illustration_of_A_and_A_star_and_B}. This table is constructed based on the layout of Table 1 in~\citet{klochowicz2025neglect}.}
  \label{table: example of items}
\end{table*}

\begin{figure*}\label{fig: pair_of_pictures_A_and_A_star_and_B}
    \centering
    \begin{minipage}[b]{0.3\linewidth}
        \centering
        \includegraphics[width=\linewidth]{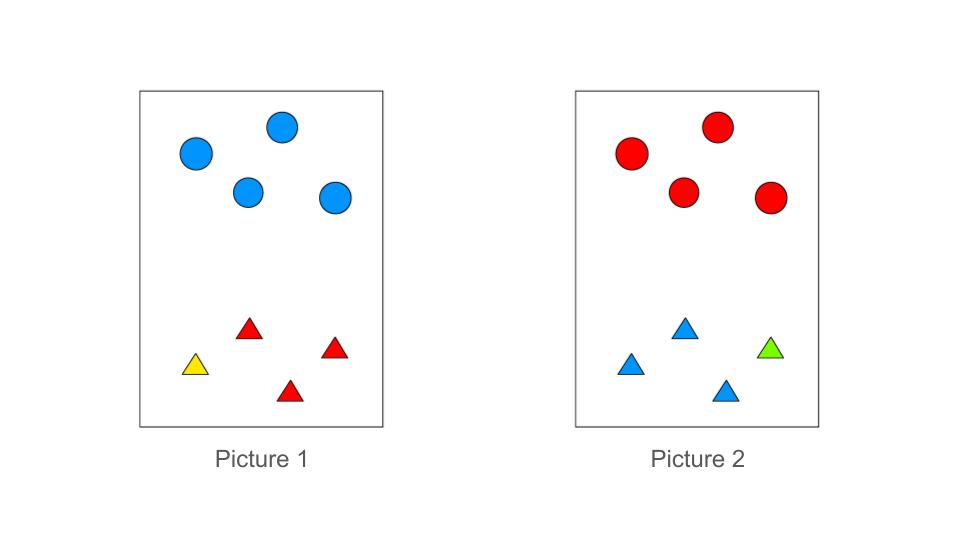}
        \subcaption{Example of A in Table~\ref{table: example of items}.}
        \label{fig: pair_of_pictures_A}
    \end{minipage}
    \begin{minipage}[b]{0.3\linewidth}
        \centering
        \includegraphics[width=\linewidth]{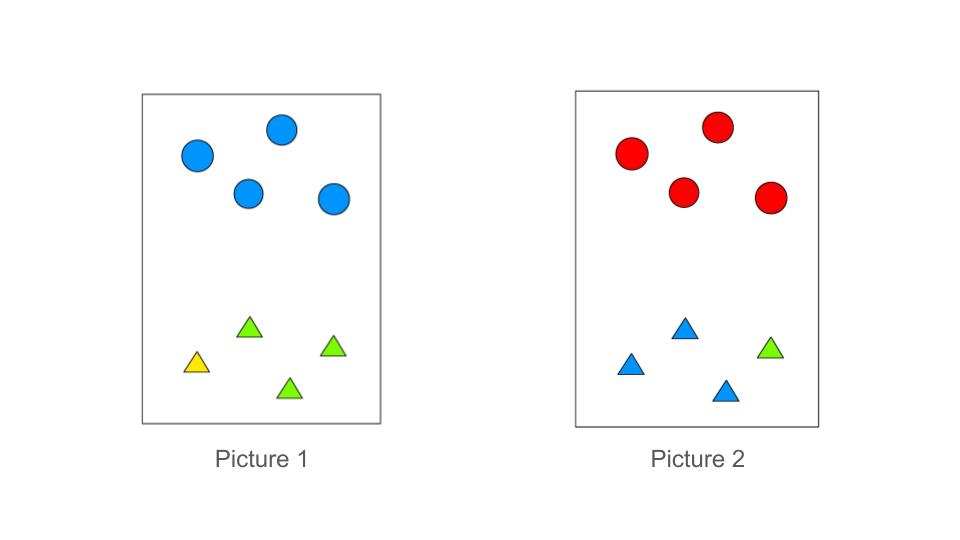}
        \subcaption{Example of A* in Table~\ref{table: example of items}.}
        \label{fig: pair_of_pictures_A_star}
    \end{minipage}
    \begin{minipage}[b]{0.3\linewidth}
        \centering
        \includegraphics[width=\linewidth]{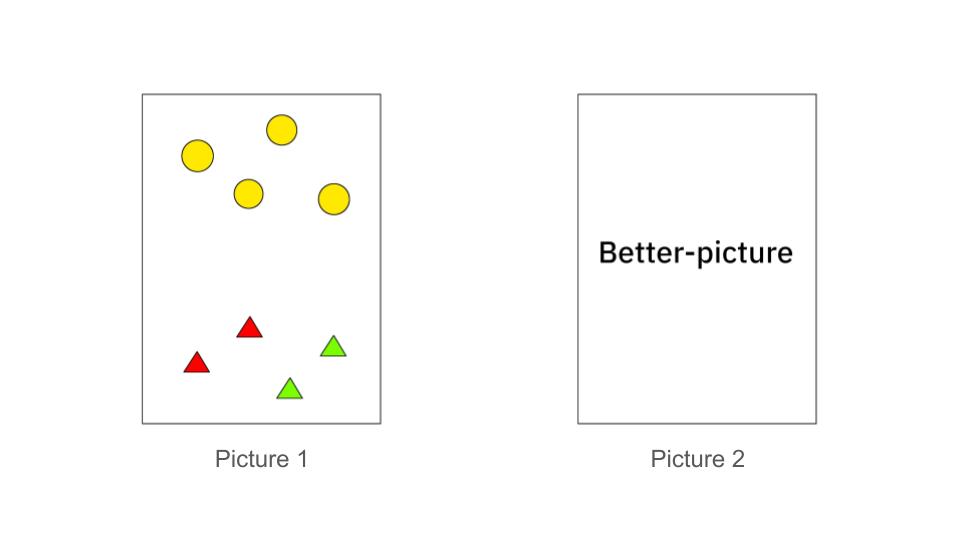}
        \subcaption{Example of B in Table~\ref{table: example of items}.}
        \label{fig: pair_of_pictures_B}
    \end{minipage}
    \caption{Illustration of A, A*, and B in Table~\ref{table: example of items}.}
    \label{fig: illustration_of_A_and_A_star_and_B}
\end{figure*}

\section{Examples of experimental items}\label{app: examples_of_task}

Examples of experimental items for each sub-experiment are shown in Table \ref{table: example of items} and Figure \ref{fig: illustration_of_A_and_A_star_and_B}.

\section{Prompts}\label{app: prompts}

The prompt format provided to the LLM in a single trial is shown below.
\begin{tcolorbox}
    Currently the following pair of pictures is presented:\\
    \{The explanation of Picture 1\}\\
    \{The explanation of Picture 2\}\\
    The sentence is: \{sentence\}\\
    Which picture does match with the sentence in this situation? Here are your answer options:\\
    \{option1\}\\
    \{option2\}\\
    Your answer: I choose
\end{tcolorbox}
\noindent
A concrete example of it is shown below.
This example uses the target row in Table~\ref{table: example of items} as the prompt.
\begin{tcolorbox}
    Currently the following pair of pictures is presented:\\
    Picture 1 contains 4 yellow circles in the upper half, and 2 red triangles and 2 green triangles in the lower half.\\
    Picture 2 is the better-picture.\\
    The sentence is: Fewer than three of the circles are red.\\
    Which picture does match with the sentence in this situation? Here are your answer options:\\
    Second\\
    First\\
    Your answer: I choose
\end{tcolorbox}
\noindent
The prompt format provided to the LLM in each experiment is shown below.
\begin{tcolorbox}[breakable=true]
    In the following, we will ask for your judgments about certain kinds of sentences in English.\\
    The sentences refers to pairs of pictures.
    The picture contains two types of geometrical shapes, one in the upper half and one in the lower half of the picture.\\
    One of these shapes in the picture were homogeneous with respect to their color, and the other set had mixed colors, containing one element with a different color.\\
    Only one of the pictures would match with the sentence in each trial, and your task is to choose that one.\\
    The covered picture, what we call "better-picture", is sometimes contained in the pairs.\\
    \\
    You will see many pairs, each of which will be accompanied by an sentence about the contents of the pictures. \\
    Your task is to decide which picture in a pair match this sentence.\\
    \\The better-picture should only be chosen if the open picture did not match the sentence.
    You will answer 'First' if you consider the picture 1 match the sentence; otherwise you will answer 'Second'.\\
    Do not include any words or sentences other than "First" or "Second" in your answer.\\
    \\
    You will start with a short training to get you familiar with the response procedure. \\
    During this training, you will see examples of correct responses.\\
    \#\#\# Training 1\\
    \{fewshot\_1\}\\
    \\
    \#\#\# Training 2\\
    \{fewshot\_2\}\\
    \\
    \#\#\# Training 3\\
    \{fewshot\_3\}\\
    \\
    \#\#\# Training 4\\
    \{fewshot\_4\}\\
    \\
    \#\#\# Training 5\\
    \{fewshot\_5\}\\
    \\
    \#\#\# Training 6\\
    \{fewshot\_6\}\\
    \\
    \#\#\# Training 7\\
    \{fewshot\_7\}\\
    \\
    \#\#\# Training 8\\
    \{fewshot\_8\}\\
    \\
    \#\#\# Your turn\\
    As in the training, you will decide which pictures are appropriate to the sentence you see.\\
    \{48 experimental items\}
\end{tcolorbox}

Regarding the task dataset before the conversion, we utilize the experimental dataset in CSV format provided by~\citet{klochowicz2025neglect} and annotate it with additional necessary data.
Specifically, we add information about Picture 2 to the data for all sub-experiments.\footnote{Information about Picture 1 is already included.}
For the filler trial data, we additionally append labels indicating which of Picture 1 or Picture 2 matches the sentence, and the validation of this annotation is performed by two annotators.
Additionally, the filler trials originally contain sentences of the form ``Half of the \{shapes\} are \{color\},'' but because the phrase ``Half of'' may trigger scalar implicature, we replace them with ``Exactly half of the \{shapes\} are \{color \}.''

\section{GLMM}\label{app: glmm}

Before explaining a GLMM, we first define several technical terms.

First, the observed data refer to the \textit{response variable}, while the data considered to affect the variation in the response variable is termed the \textit{explanatory variable}.
The explanatory variables include the priming condition, the trial index, the intercepts and slopes for each seed, and the intercepts for each experimental item.

Next, the effects considered to affect the observed data are categorized into \textit{fixed effects} and \textit{random effects}.
Fixed effects refer to the factors of primary interest in the analysis.
Random effects refer to factors that are not the primary focus of the analysis but are expected to influence the observed data.

The GLMM equation used in this experiment is restated below.
\begin{align*}
    \text{logit}(q_i) = \beta_i + \sum_{j}{\beta_{ji}x_{ji}} + \sum_k{r_{ki}}.
\end{align*}
It is assumed that the response variable follows a binomial distribution, and that the random effects $r_{kj}$ independently follow a normal distribution.
The analysis is performed by estimating the model parameters using maximum likelihood estimation and conducting statistical tests.

\section{Excluded Results}\label{app: human_results}

The mean accuracy in filler trials of Gemma-3-1B, Gemma-3-4B, and Gemma-3-12B is shown in Table~\ref{table: analysis_gpt_excluded}.
\begin{table}[tbp]
  \centering
  \small
    \begin{tabular}{lllll}
      \toprule
       & Gemma-3-1B & Gemma-3-4B & Gemma-3-12B \\
      \midrule
      ESQ & 23.8 & 43.7 & 59.3 \\
      DIS & 24.2 & 40.2 & 59.2 \\
      UPP & 25.9 & 40.3 & 62.0 \\
      BAS & 25.9 & 49.7 & 62.0 \\
      \bottomrule
    \end{tabular}
    \caption{Summary of the mean accuracy in filler trials of Gemma-3-1B, Gemma-3-4B, and Gemma-3-12B. Their seeds were excluded because of their low accuracy in filler trials.}
  \label{table: analysis_gpt_excluded}
\end{table}

\end{document}